\documentclass{article} 
\usepackage{hyperref}
\usepackage{graphics}
\usepackage{amsthm}
\usepackage{mathtools}
\usepackage{amsmath,amsfonts,amssymb}       
\usepackage{multirow}
\usepackage{multicol}
\usepackage{floatrow}
\usepackage{appendix}
\usepackage{bbm}
\usepackage{graphicx}
\usepackage{color}

\theoremstyle{definition}

\theoremstyle{definition}

\usepackage{appendix}
\usepackage{authblk}
\usepackage{algpseudocode}
\usepackage{algorithm}
\usepackage[letterpaper,margin=1.75in]{geometry}
\usepackage[font={small,it}]{caption}
\graphicspath{{./figures/}}
\usepackage{expl3}[2012-07-08]
\ExplSyntaxOn
\cs_new_eq:NN \fpeval \fp_eval:n
\ExplSyntaxOff

\title{Variational Composite Autoencoders}
\author[1,2]{Jiangchao Yao}
\author[2]{Ivor Tsang}
\author[1]{Ya Zhang}
\affil[1]{Shanghai Jiao Tong University}
\affil[2]{University of Technology Sydney}

\date{}

\begin{document}
\maketitle

\begin{abstract}
Learning in the latent variable model is challenging in the presence of the complex data structure or the intractable latent variable. Previous variational autoencoders can be low effective due to the straightforward encoder-decoder structure. In this paper, we propose a variational composite autoencoder to sidestep this issue by amortizing on top of the hierarchical latent variable model. The experimental results confirm the advantages of our model.
\end{abstract}

\section{Introduction}
The latent variable model in Figure~\ref{fig:mm}(a) has been widely explored for the representation learning. One line of research has concentrated on a broad range of data structures. For example, \cite{kipf2016variational} proposed a variational graph autoencoder to consider the graph-structured data. \cite{chen2016variational} introduced a variational lossy autoencoder to learn the controllable representation of the image. \cite{kusner2017grammar} proposed a grammar variational autoencoder to directly learn from a parse tree described by the grammar. Another line of research pays attention to the optimization difficulty on the intractable latent variable. For instance, \cite{mnih2014neural} proposed a general score function method with control variates to optimize the model with continuous or discrete latent variables. \cite{maddison2016concrete} introduced a \emph{Concrete distribution}, the continuous relaxation of the discrete distribution, to approximately reparameterize the discrete latent variable. \cite{nalisnick2016stick} extended variational autoencoders to perform posterior inference for the latent variable of Stick-Breaking processes.

Although the promising performance has been achieved, previous variational autoencoders in Figure~\ref{fig:mm}(c) can be low effective due to the straightforward encoder-decoder structure. In the presence of the very sophisticated data or the unstable optimization on the latent variable, it is challenging with only resort to the implementation of the encoder and the decoder. This problem motivates us to explore a more expressive structure in the representation learning, which can further improve the encoding efficiency and ease the optimization burden. 

In this paper, we propose a variational composite autoencoder (VCAE) to sidestep this issue by amortizing on top of the hierarchical latent variable model. As shown in Figure~\ref{fig:mm}(b), a surrogate variable is introduced between the latent variable and the data to intermediate their gap. Specifically, in terms of the complex data structure, the representation learning is implemented hierarchically, i.e., the raw data is first projected into a moderate surrogate space, on top of which a latent variable is then learned. For the optimization difficulty on the latent variable, VCAE provides a possible framework to unify previous approaches, score function methods and reparameterization. Experimental results demonstrate the superior performance of VCAE over state-of-the-arts in these two aspects.


\section{Variational Composite autoencoders}
\begin{figure*}[t]
\centering
\includegraphics[width=0.8\textwidth]{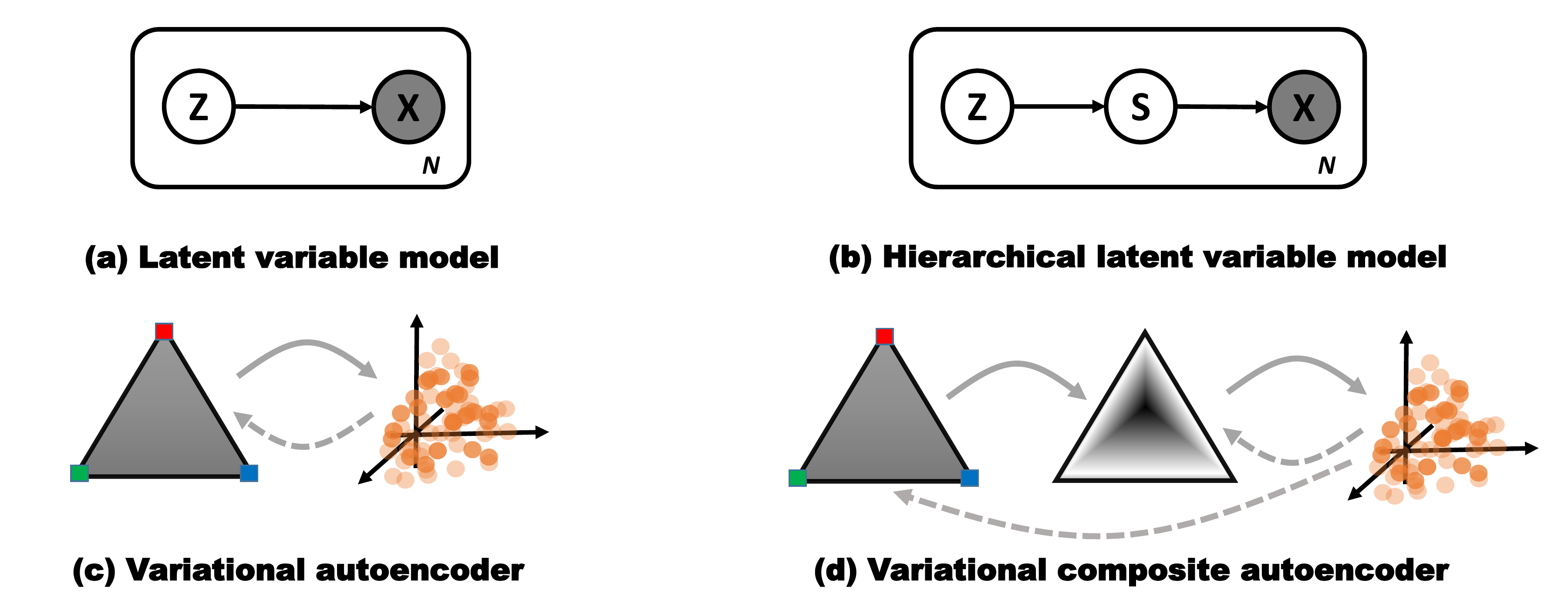}
\caption{The idea of our work. (a) and (b) are the graphical representations of two models. $z$ represents the latent variable, $s$ is the surrogate variable and $x$ is the observation. (c) and (d) illustrate the variational autoencoder and our variational composite autoencoder. We introduce a surrogate variable to sidestep the learning difficulty between the latent variable and the observation.}
\label{fig:mm}
\end{figure*}

The key idea is illustrated in Figure~\ref{fig:mm}. Compared with the one-off encoding and decoding process in variational autoencoder (VAE) \cite{kingma2013auto}, we decouple the representation learning into two sub-procedures by introducing a surrogate variable. Then the learning difficulty either on the sophisticated data structure or the unstable optimization can be alleviated in this surrogate space. 

In VCAE, we use two inference networks $q_{\theta_1}(s|x)$ and $q_{\theta_2}(z|x)$ to represent the encoding processes from $x$ to $s$ and $z$ respectively. The generative processes from $z$ to $s$ and $s$ to $x$ are modeled with two networks $p_\psi(s|z)$ and $p_\phi(x|s)$. The graphical representation of our hierarchical latent variable model is illustrated in Figure~\ref{fig:mm}(b).
According to the encoding and decoding processes in Figure~\ref{fig:mm}(d), we can deduce a variational lower bound as follows,
\begin{align} \label{eq:model}
\ln{p(x)} \geq \mathbb{E}_{q_{\theta_1}(s|x)}\left[\ln{p(x|s)}\right] - \mathbb{E}_{q_{\theta_2}(z|x)}\left[\text{D}_{KL}(q_{\theta_1}(s|x)||p_\psi(s|z))\right] - \text{D}_{KL}(q_{\theta_2}(z|x)||p(z)). 
\end{align}
Previous stochastic optimization on the variational Monte Carlo objective can be seamlessly applied to optimize this bound, which makes VCAE general to many applications. For example, when both $z$ and $s$ are reparameterizable, the objective can be directly optimized by the stochastic backpropagation \cite{rezende2014stochastic}. When $z$ is the discrete variable and $s$ is chosen to be reparameterizable, score function methods can be used jointly with reparameterization for optimization. In this paper, we will focus on the comparison between VAE and VCAE in the representation learning, and the optimization effectiveness of VCAE compared with previous state-of-the-arts. 

\section{Evaluation}
We evaluate the effectiveness of VCAE using two benchmark tasks, generative modeling and structured prediction, with the statically binarized MNIST dataset \cite{salakhutdinov2008quantitative}. The standard partition 50000/10000/10000 is used to split the MNIST dataset into the training, validation and test sets. All sub-modules in VCAE are parameterized with the sigmoid belief network. Specifically, 200 stochastic units interleaved by one linear layer (denoted as \textbf{Linear}) or two layers of 200 tanh units (denoted as \textbf{Nonlinear}) are used for two encoding channels $q_{\theta_1}(s|x)$, $q_{\theta_2}(z|x)$ and one decoding channel $p_\phi(x|s)$. One sigmoid layer is fixed for $p_\psi(s|z)$. All models are randomly initialized and optimized by the ADAM optimizer \cite{kingma2014adam} with a learning rate of 3e-4. During the test, we compute the importance-sampled estimate of the log-likelihood objective.  

\begin{figure}
    \centering
    \begin{tabular}{cc}
    \!\!\!\! \!\!\!\! \!\!\!\! \includegraphics[width=0.45\linewidth]{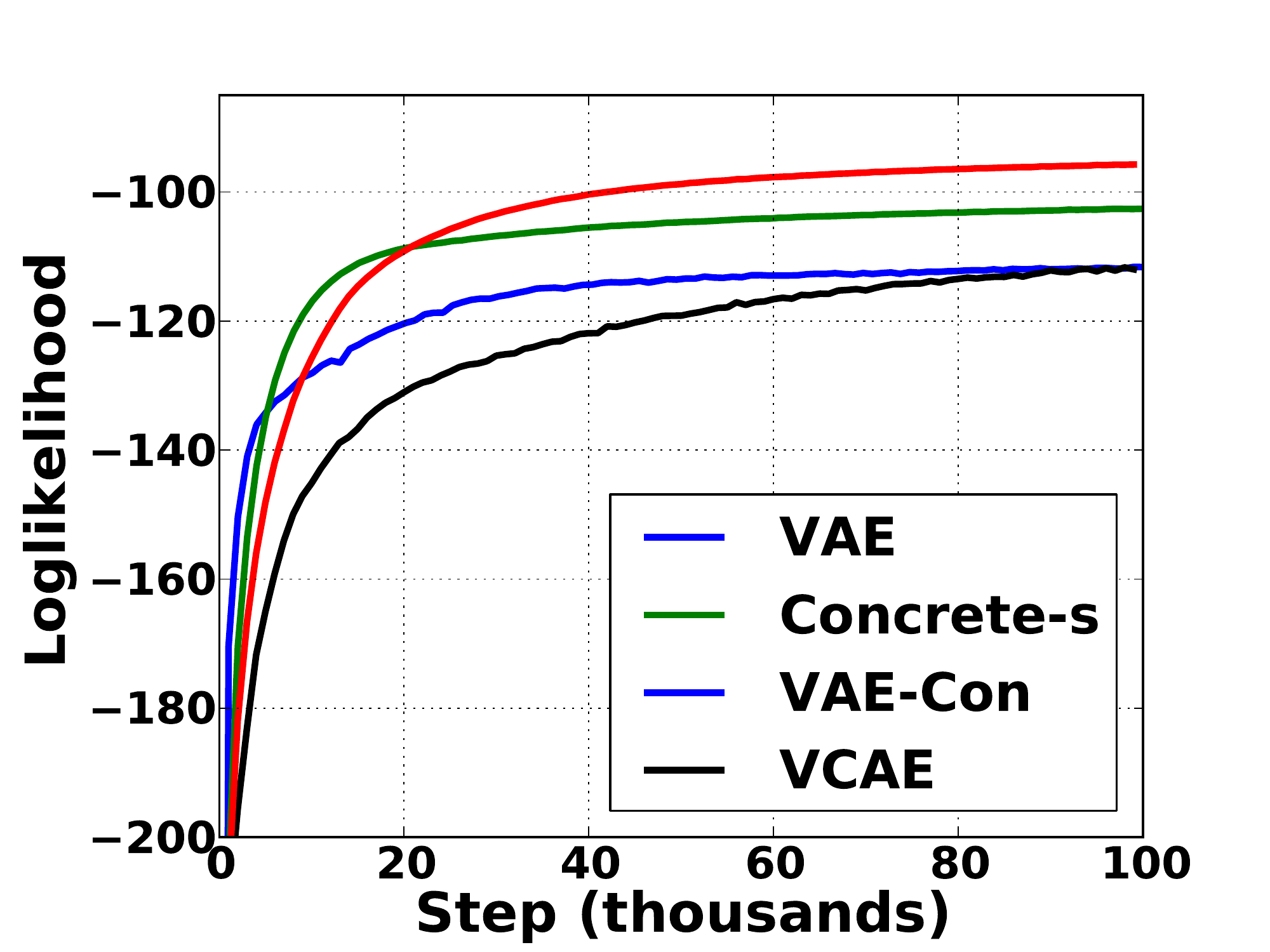} &  \!\!\!\!  \!\!\!\! 
    \includegraphics[width=0.45\linewidth]{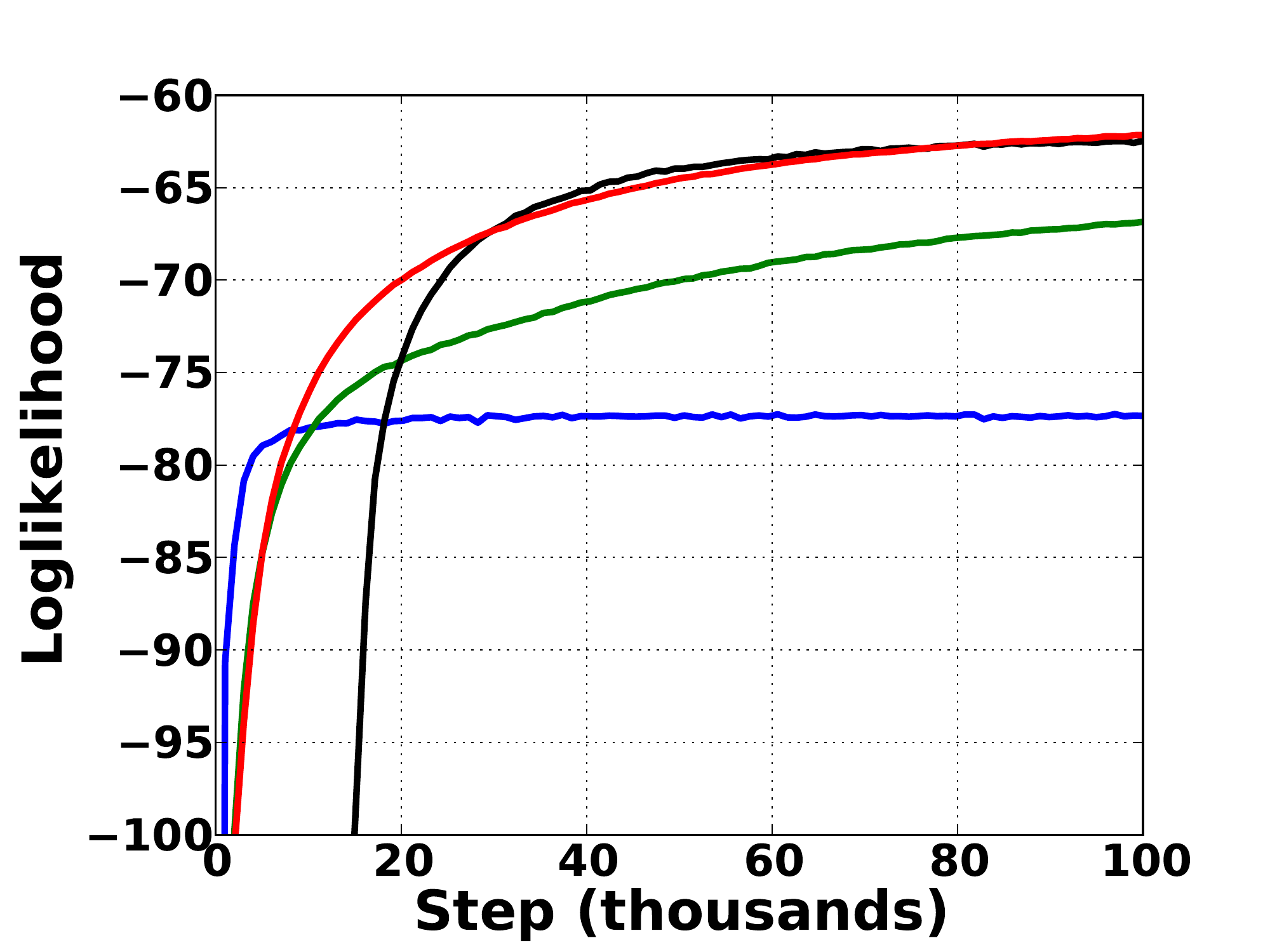}
	\end{tabular}
    \captionof{figure}{\small Comparison among VAE, Concrete-s, the sequential combination of VAE and Concrete-s (VAE-Con), and VCAE on generative modeling (left) and structured prediction (right).}
\label{mnist_sbn_sp_vae}
\end{figure}

\begin{table}
\centering
\renewcommand{\arraystretch}{1.1}
\scalebox{0.9}{\begin{tabular}{c|cc}
\multirow{2}{*}{Methods} & Generative & Structured \\
& modeling & prediction \\
\hline
\multirow{2}{*}{VAE} & \multirow{2}{*}{-112.8} & \multirow{2}{*}{-77.1} \\
 & & \\
\multirow{2}{*}{Concrete-s} & \multirow{2}{*}{-105.3} & \multirow{2}{*}{-66.3} \\
& & \\
\multirow{2}{*}{VAE-Con} & \multirow{2}{*}{-111.5} & \multirow{2}{*}{-62.4} \\
 & & \\
\multirow{2}{*}{VCAE} & \multirow{2}{*}{\textbf{-95.7}} & \multirow{2}{*}{\textbf{-62.1}} \\
\end{tabular}
}
\captionof{table}{\small The importance-weighted estimate of the log-likelihood objective on two tasks in the test dataset.}
\label{tab:results_gm_sp_mnist_test0}
\end{table}

In the first experiment, we use VCAE to directly compare with VAE~\cite{kingma2013auto}. In the hierarchical latent variable model, the Gaussian variable and the reparameterizable Concrete variable~\cite{maddison2016concrete} are respectively chosen as the latent variable and the surrogate variable. Thus our VCAE will includes the Gaussian reparameterization (VAE) and the Concrete reparameterization (Concrete-s) parts. To comprehensively understand our method, we also use Concrete-s and the sequential combination (VAE-Con) of VAE and Concrete-s as two baselines in the experiments. Figure~\ref{mnist_sbn_sp_vae} and Table~\ref{tab:results_gm_sp_mnist_test0} present the performance of three methods in the Linear case. As can be seen, VCAE outperforms two baselines in the training procedure, and achieves $-95.7$ and $-62.1$ on the two tasks respectively, which is significantly better than VAE. Besides, as an important component in our method, Concrete-s also shows promising performance, but still underperforms VCAE. This indicates our hierarchical representation learning structure is more effective. 

\begin{table*}
\caption{\small The importance-weighted estimate of the log-likelihood objective in the test dataset.}
\label{tab:results_gm_sp_mnist_test1}
\renewcommand{\arraystretch}{1.2}
\begin{center}
\scalebox{0.95}{
\begin{tabular}{c|c|ccc|c|c }
\multicolumn{2}{c|}{Methods}  & NVIL & REBAR & Concrete-z & Concrete-s & VCAE\\
\hline
Generative & Linear & -108.3 & -107.6 & -107.5 & -105.3 & \textbf{-97.9} \\
modeling & Nonlinear & -100.9 & -102.0 & -101.2 & -100.6 & \textbf{-95.0} \\
\hline
Structured & Linear & -67.9 & -66.4 & -66.1 & -66.3 & \textbf{-64.4} \\
prediction & Nonlinear & -63.8 &  -63.5 & -67.0 & -63.3 & \textbf{-62.9} \\
\end{tabular}
}
\end{center}
\end{table*}

\begin{figure*}
\begin{center}
\begin{tabular}{cccc}
    \!\!\!\!\includegraphics[width=0.22\linewidth]{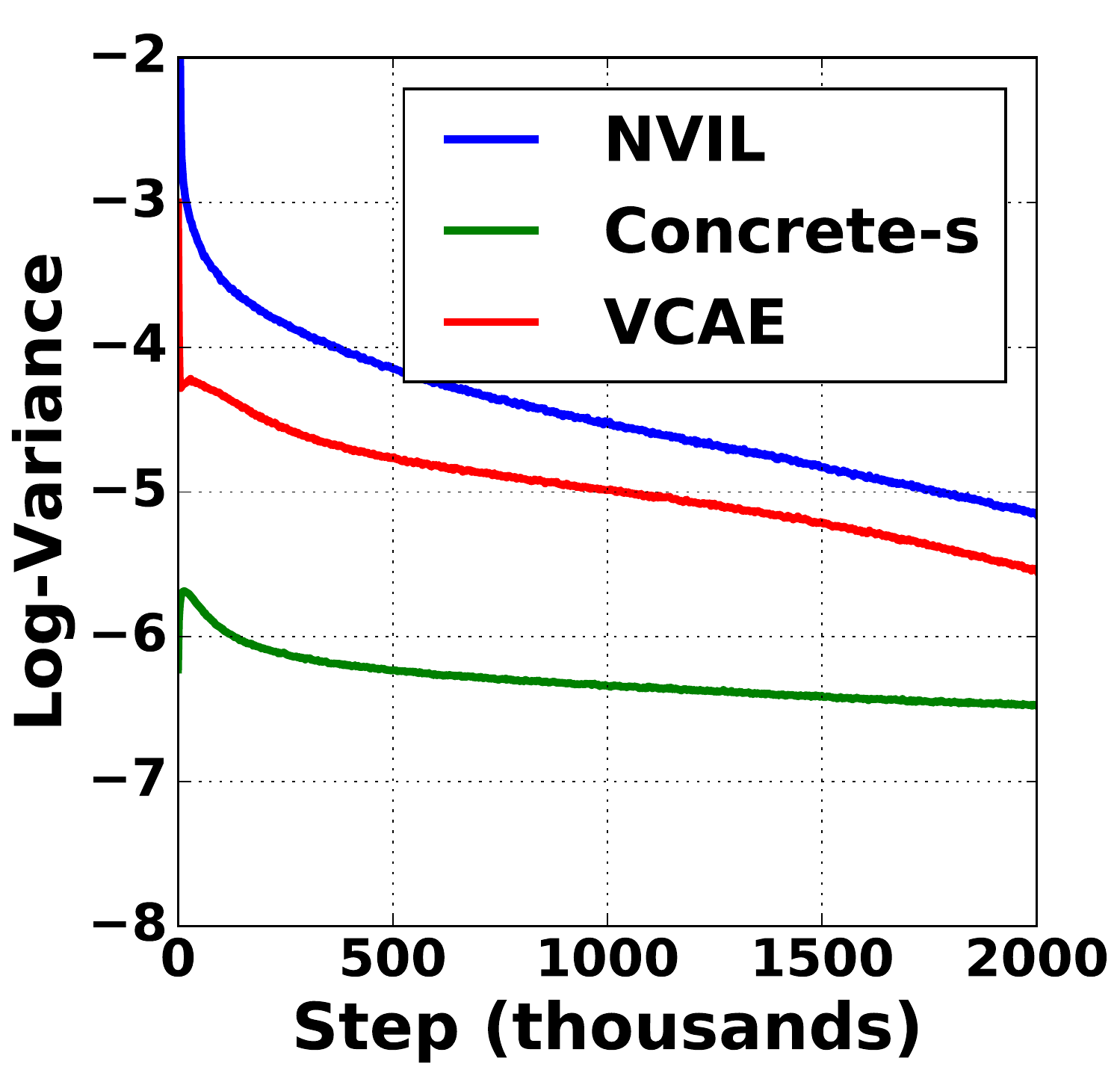} \!\!\!\! & \!\!\!\!
    \includegraphics[width=0.22\linewidth]{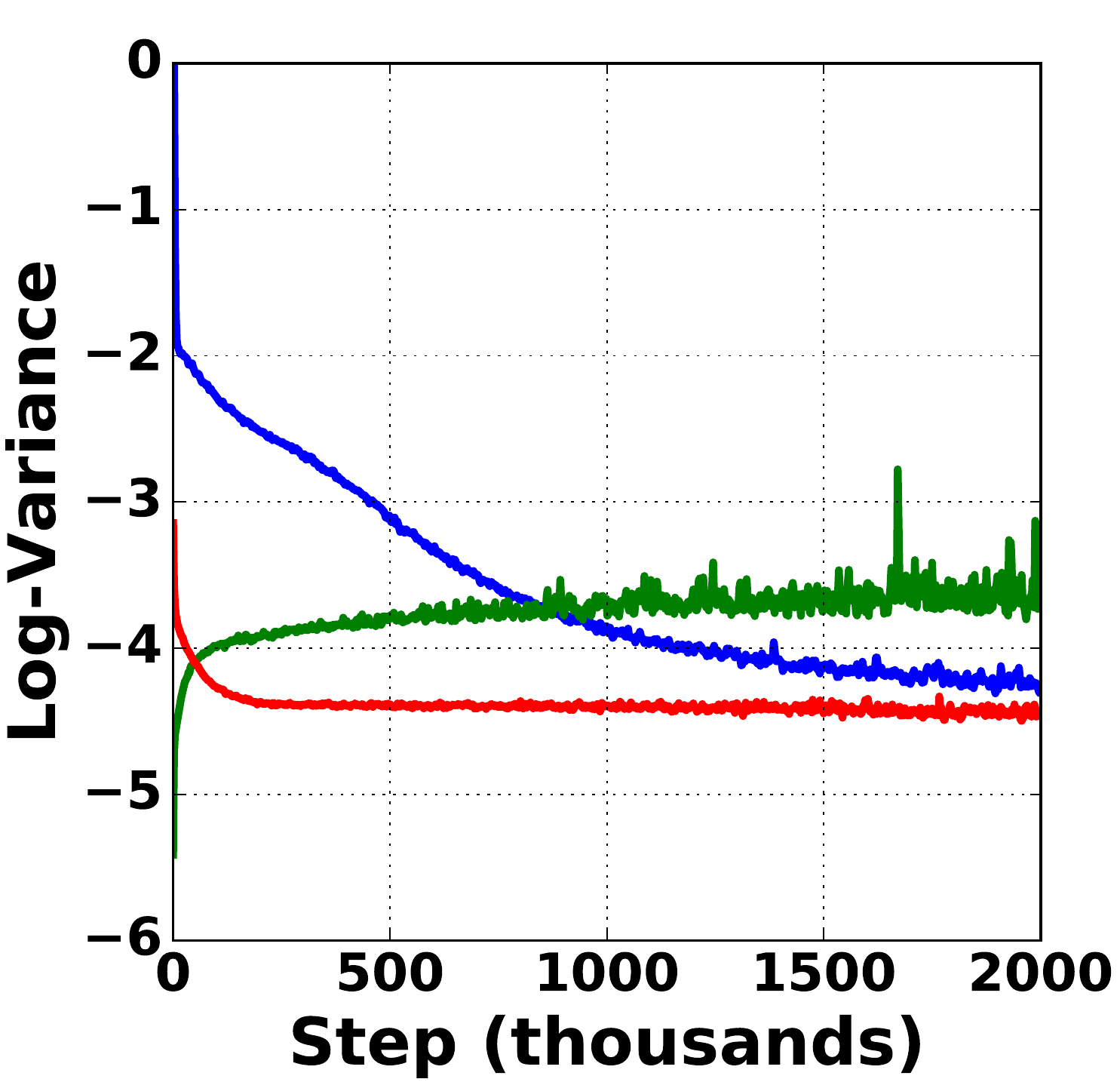} \!\!\!\! & \!\!\!\!
    \includegraphics[width=0.22\linewidth]{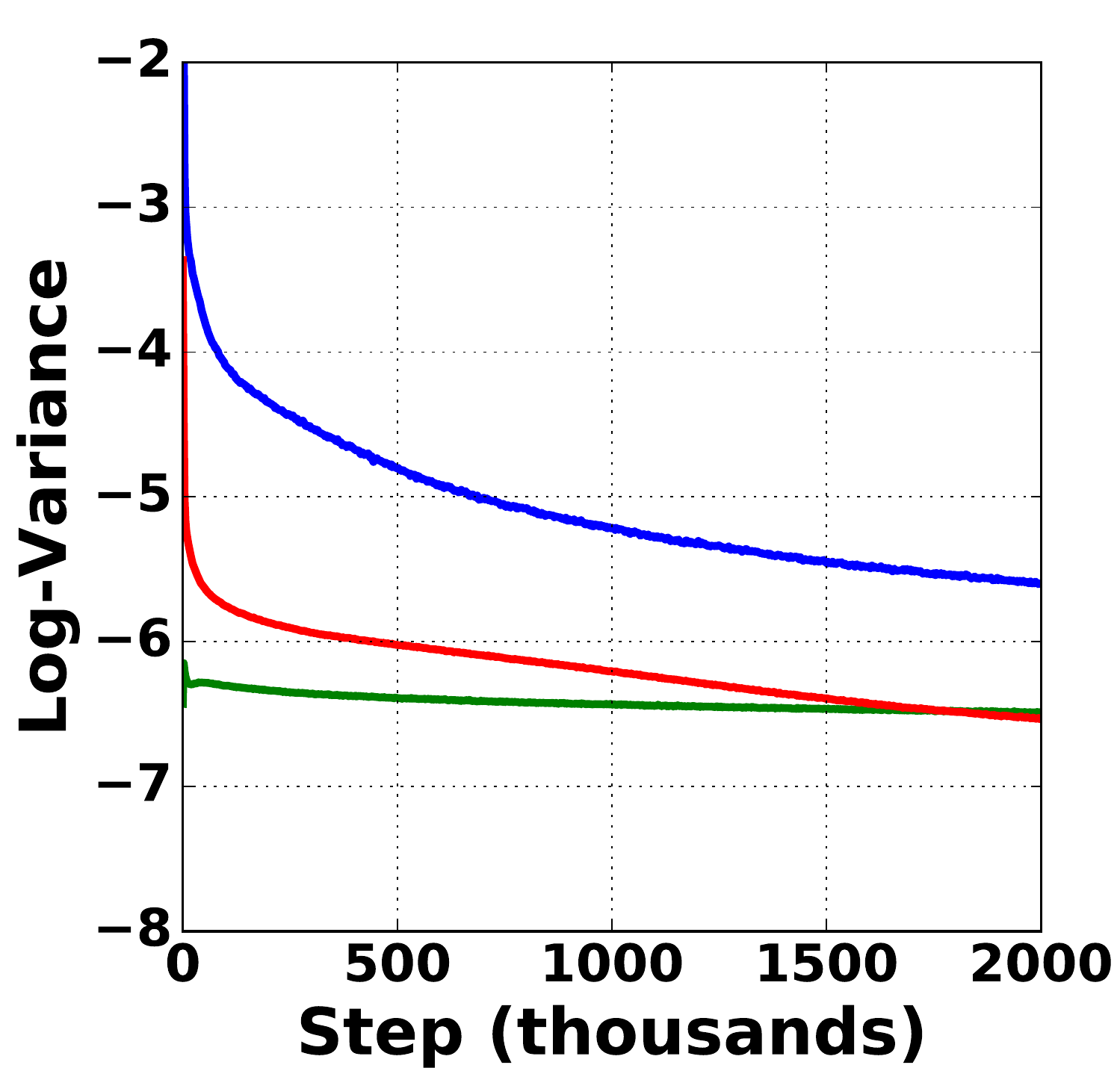} \!\!\!\! & \!\!\!\!
   \includegraphics[width=0.22\linewidth]{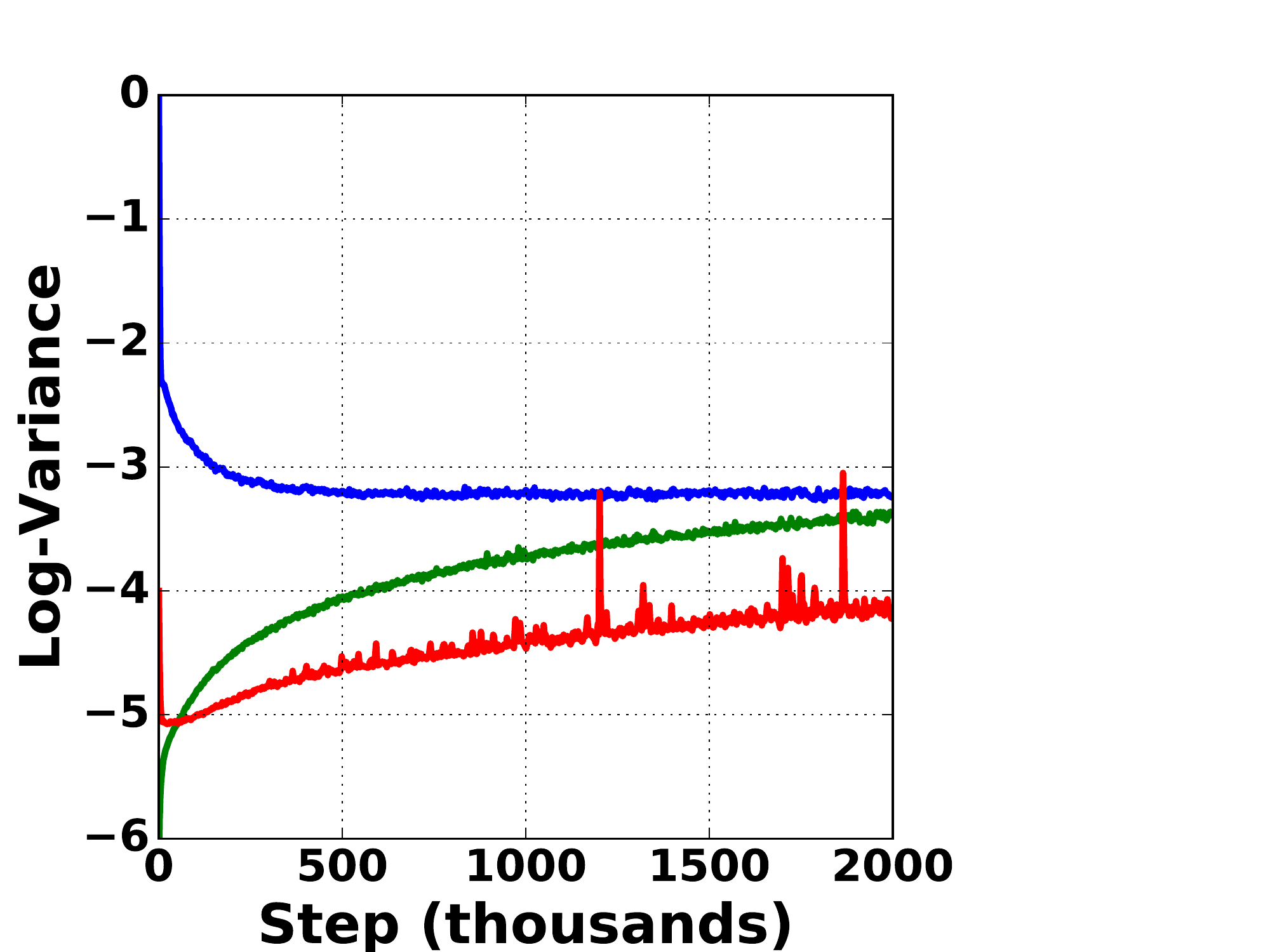}
\end{tabular}
\caption{\small Gradient variance comparison among NVIL, Concrete-s and VCAE (Linear: the 1st and 3rd figures, Nonlinear: the 2nd and 4th figures) on generative modeling (left two) and structured prediction (right two).}
\label{mnist_variance_inter}
\end{center}
\vskip -0.2in
\end{figure*}

In the second experiment, we use VCAE to optimize the challenging discrete variable model. Similarly, the Concrete variable is set as the surrogate variable. Then we unify the score function method, NVIL \cite{mnih2014neural}, and the reparamterization method, Concrete-s, into VCAE by applying them on $q_{\theta_2}(z|x)$ and $q_{\theta_2}(s|x)$ respectively. The resulting method is to compare with the state-of-the-art approaches and the final performance is summarized in Table~\ref{tab:results_gm_sp_mnist_test1}. For the generative modeling task, the importance-weighted estimate of the log-likelihood objective for VCAE (Linear) is $-97.9$, while that of the best unbiased estimators REBAR (Linear) \cite{Tucker2017REBARLU} and the biased reparameterization method Concrete-z\footnote{In the test phase, the binarized Concrete samples are used to evaluate the lower bounds.} are $-107.6$ and $-107.5$, respectively.  For the structured prediction task, VCAE outperforms all unbiased estimators and the biased reparameterization method (Concrete-z) even in the case of continuous variable (Concrete-s). This confirm the advantages of VCAE in the optimization. To present a fine-grained performance analysis of VCAE, we further compare VCAE with NVIL and Concrete-s by tracing their gradient variance in the training phase, as the lower gradient variance the faster convergence is~\cite{johnson2013accelerating}. As shown in Figure~\ref{mnist_variance_inter}, in the Linear case, the gradient variance of VCAE is between that of NVIL and Concrete-s, and  in the Nonlinear case, VCAE not only outperforms NVIL, but also surpasses Concrete-s. This indicates  on the basis of the score function method in Eq.~\eqref{eq:model}, the variance is further reduced due to the reparameterizable Concrete variable, and the improvement is even better than both of them.  

\section{Conclusion}
In this paper, we propose a variational composite autoencoder to improve the effectiveness of representation learning. Compared with previous VAEs, VCAE enjoys the advantages of amortizing on top of the hierarchical latent variable model. By introducing a surrogate variable to learn the intermediate space, both learning burden and optimization difficulty can be alleviated. Experimental results in these two aspects demonstrate VCAE outperforms the state-of-the-art methods.

\bibliography{ref}

\begin{thebibliography}{10}

\bibitem{chen2016variational}
Xi~Chen, Diederik~P Kingma, Tim Salimans, Yan Duan, Prafulla Dhariwal, John
  Schulman, Ilya Sutskever, and Pieter Abbeel.
\newblock Variational lossy autoencoder.
\newblock {\em ICLR}, 2017.

\bibitem{johnson2013accelerating}
Rie Johnson and Tong Zhang.
\newblock Accelerating stochastic gradient descent using predictive variance
  reduction.
\newblock In {\em Advances in neural information processing systems}, 2013.

\bibitem{kingma2014adam}
Diederik Kingma and Jimmy Ba.
\newblock Adam: A method for stochastic optimization.
\newblock {\em ICLR}, 2015.

\bibitem{kingma2013auto}
Diederik~P Kingma and Max Welling.
\newblock Auto-encoding variational bayes.
\newblock {\em ICLR}, 2014.

\bibitem{kipf2016variational}
Thomas~N Kipf and Max Welling.
\newblock Variational graph auto-encoders.
\newblock {\em arXiv preprint arXiv:1611.07308}, 2016.

\bibitem{kusner2017grammar}
Matt~J Kusner, Brooks Paige, and Jos{\'e}~Miguel Hern{\'a}ndez-Lobato.
\newblock Grammar variational autoencoder.
\newblock {\em Proceedings of the 34th International Conference on Machine
  Learning}, 2017.

\bibitem{maddison2016concrete}
Chris~J Maddison, Andriy Mnih, and Yee~Whye Teh.
\newblock The concrete distribution: A continuous relaxation of discrete random
  variables.
\newblock {\em ICLR}, 2017.

\bibitem{mnih2014neural}
Andriy Mnih and Karol Gregor.
\newblock Neural variational inference and learning in belief networks.
\newblock In {\em Proceedings of the 31st International Conference on Machine
  Learning}, 2014.

\bibitem{nalisnick2016stick}
Eric Nalisnick and Padhraic Smyth.
\newblock Stick-breaking variational autoencoders.
\newblock {\em ICLR}, 2017.

\bibitem{rezende2014stochastic}
Danilo~Jimenez Rezende, Shakir Mohamed, and Daan Wierstra.
\newblock Stochastic backpropagation and approximate inference in deep
  generative models.
\newblock {\em Proceedings of the 31st International Coference on International
  Conference on Machine Learning}, 2014.

\bibitem{salakhutdinov2008quantitative}
Ruslan Salakhutdinov and Iain Murray.
\newblock On the quantitative analysis of deep belief networks.
\newblock In {\em Proceedings of the 25th international conference on Machine
  learning}, 2008.

\bibitem{Tucker2017REBARLU}
George Tucker, Andriy Mnih, Chris~J. Maddison, and Jascha Sohl-Dickstein.
\newblock Rebar: Low-variance, unbiased gradient estimates for discrete latent
  variable models.
\newblock {\em Advances in Neural Information Processing Systems}, 2017.

\end{thebibliography}
\bibliographystyle{plain}

\end{document}